\documentclass[lettersize,journal]{IEEEtran}
\usepackage{amsmath,amsfonts}
\usepackage{algorithmic}
\usepackage{array}
\usepackage{textcomp}
\usepackage{stfloats}
\usepackage{subfigure}
\usepackage{url}
\usepackage{verbatim}
\usepackage{graphicx}
\def\BibTeX{{\rm B\kern-.05em{\sc i\kern-.025em b}\kern-.08em
    T\kern-.1667em\lower.7ex\hbox{E}\kern-.125emX}}
\usepackage{balance}
\usepackage{multirow}
\usepackage{cite}
\usepackage{color}
\usepackage{ulem}
\usepackage{chngcntr}

\begin{document}
\title{Instance-incremental Scene Graph Generation from Real-world Point Clouds via Normalizing Flows}
\author{Chao~Qi,
        Jianqin~Yin,
        Jinghang~Xu,
        and~Pengxiang~Ding
\thanks{

This work was supported partly by the National Natural Science Foundation of China under Grants 62173045 and 61673192, partly by the Fundamental Research Funds for the Central Universities under Grant 2020XD-A04-2, and partly by the BUPT Excellent Ph.D. Students Foundation under Grant CX2021222. \emph{(Corresponding author: Jianqin Yin.)}

Chao Qi is with the School of Artificial Intelligence, Beijing
University of Posts and Telecommunications, Beijing 100876, China, and also with the Standard and Metrology Research Institute, China Academy of Railway Sciences Corporation Limited, Beijing 100081, China. (e-mail: qichao@bupt.edu.cn)

Jianqin Yin, Jinghang Xu, and Pengxiang Ding are with the School of Artificial Intelligence, Beijing
University of Posts and Telecommunications, Beijing 100876, China. (e-mail: jqyin@bupt.edu.cn; xjh\_amber@bupt.edu.cn; dingpx2015@bupt.edu.cn)

}}


\maketitle

\begin{abstract}
This work introduces a new task of instance-incremental scene graph generation: Given a scene of the point cloud, representing it as a graph and automatically increasing novel instances. A graph denoting the object layout of the scene is finally generated. It is an important task since it helps to guide the insertion of novel 3D  objects into a real-world scene in vision-based applications like augmented reality. It is also challenging because the complexity of the real-world point cloud brings difficulties in learning object layout experiences from the observation data (non-empty rooms with labeled semantics). We model this task as a conditional generation problem and propose a 3D autoregressive framework based on normalizing flows (3D-ANF) to address it. First, we represent the point cloud as a graph by extracting the label semantics and contextual relationships. Next, a model based on normalizing flows is introduced to map the conditional generation of graphic elements into the Gaussian process. The mapping is invertible. Thus, the real-world experiences represented in the observation data can be modeled in the training phase, and novel instances can be autoregressively generated based on the Gaussian process in the testing phase. To evaluate the performance of our method sufficiently, we implement this new task on the indoor benchmark dataset 3DSSG-O27R16 and our newly proposed graphical dataset of outdoor scenes GPL3D. Experiments show that our method generates reliable novel graphs from the real-world point cloud and achieves state-of-the-art performance on the datasets.
\end{abstract}


\section{Introduction}

\IEEEPARstart{W}{ith} the rapid development of 3D laser scanning, the point cloud is becoming a mainstream method of real-world scene modeling \cite{LimSP22,ZhouXMMH21,RN206-Rate-Distortion} in augmented reality applications, etc. These applications often need to automatically insert novel 3D objects into real-world scenes \cite{FarbizCWZKPBK05,HanK19,GalSOK14}. Making a layout scheme to represent what to place in the scenes and the spatial relationships between the newly placed objects is necessary. This problem has not been explored. Thus, this paper proposes a novel task to address this problem, generating scene graphs to denote suitable layouts for given point cloud scenes.

\begin{figure*}[htbp]
    \centering
    \subfigure[]{
    \includegraphics[width=0.2\textwidth]{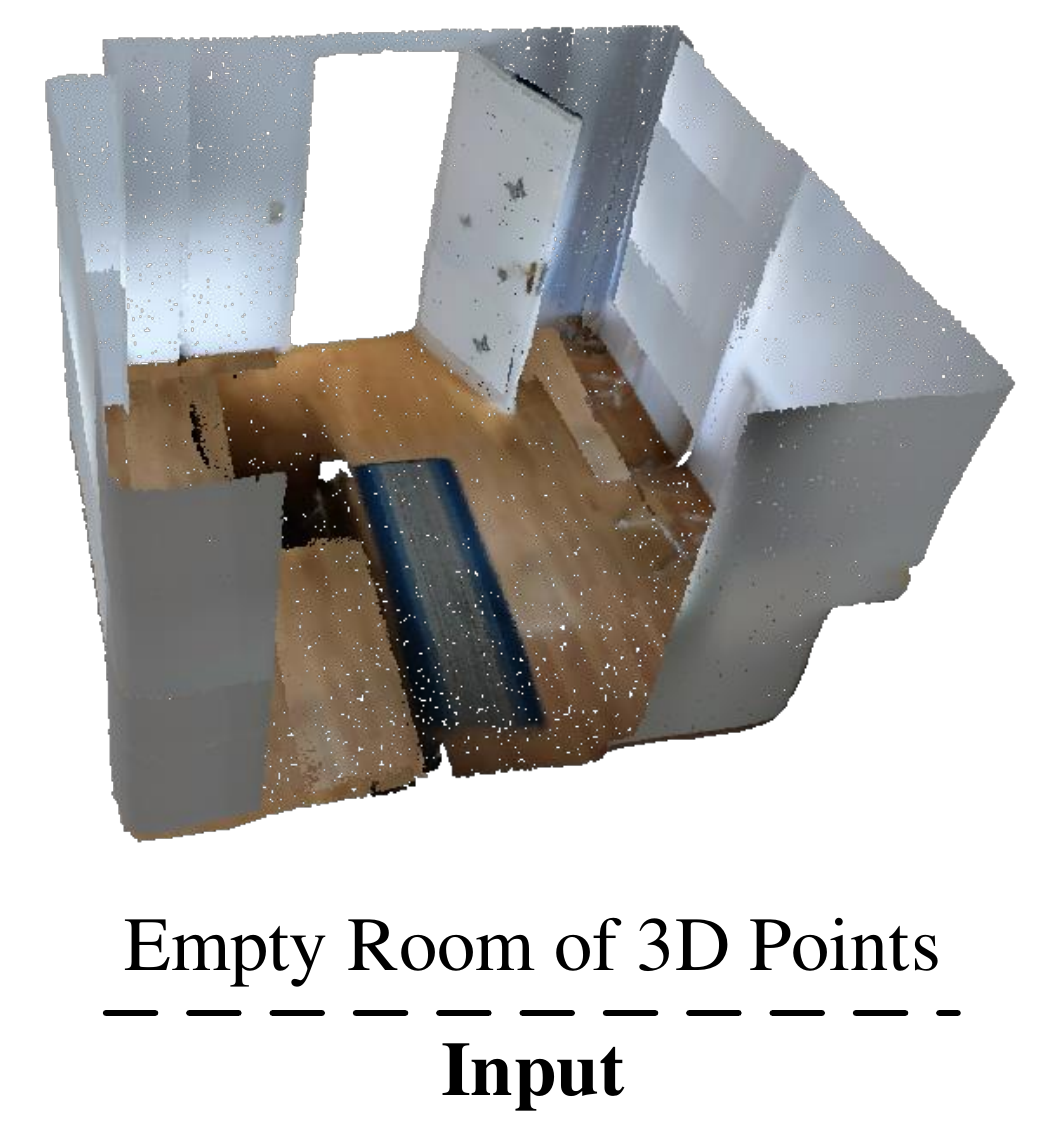}
    }
    \centering
    \subfigure[]{
    \includegraphics[width=0.73\textwidth]{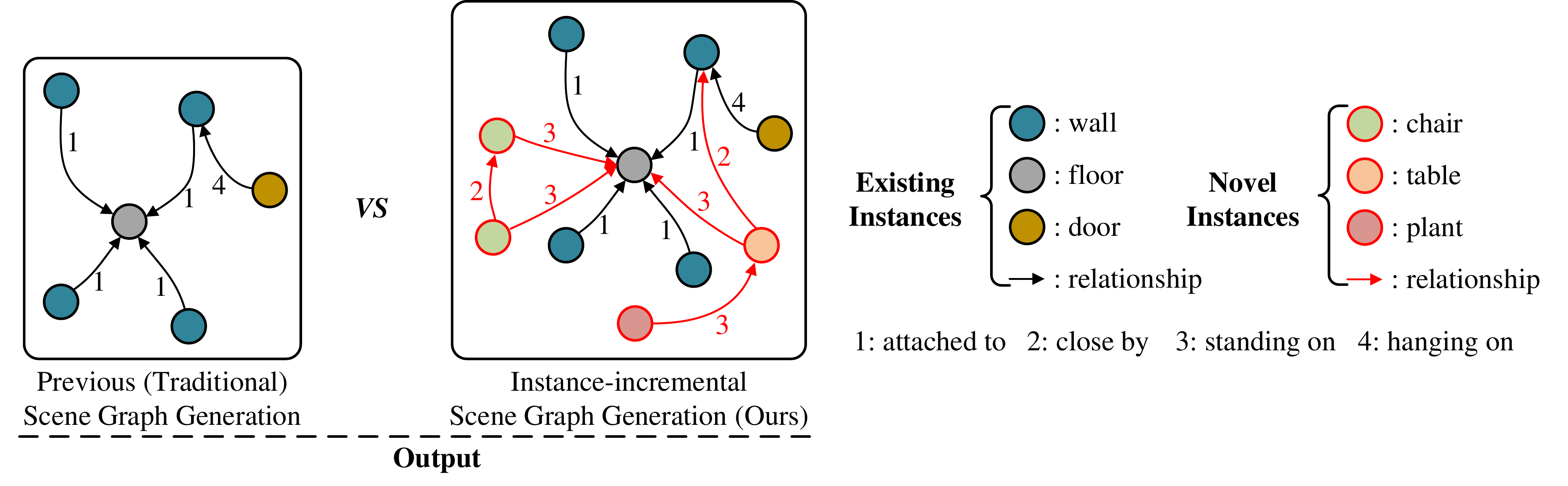}
    }
    \caption{Comparison between the previous scene graph generation and our instance-incremental generation. (a) An empty room of 3D points containing four \emph{walls}, one \emph{floor}, and one \emph{door}. (b) The previous generation task outputs a graph denoting the semantics of the six instances and their relationships. Our instance-incremental task adds novel instances (\emph{table}, \emph{chair}, etc.) to the graphical scene and establishes their relationships (e.g., `a table \emph{standing on} the floor.').}
    \label{Fig.1}
\end{figure*}

Previous 3D point-based scene graph generation \cite{ZhangY0021} focuses on analyzing scene patterns and outputting the relational semantics of existing instances in the scene. Differently, our new task aims at creating new instances and establishing the relationships between the new and the existing instances. It is a process of instance increment, and we call this new task \textbf{instance-incremental scene graph generation} to distinguish it from the previous one better. Fig . \ref{Fig.1} shows the comparison between the previous task and our new task. Taking an empty room of the point cloud as input, which contains \emph{m} existing instances, the previous task represents the given room as a graph to denote the semantics of the \emph{m} instances and their relationships \cite{RN212-Vision-Enhanced, RN211-Divide-and-Conquer, XuZCF17,YangLLBP18,LuKBL16,JohnsonGF18, LiJ19, SongCWJ22, RN210-Scene, RN208-Semantically}. Differently, Our instance-incremental generation task outputs the graph representation of \emph{m}+\emph{k} instances, in which `\emph{k}' represents the \emph{k} increased novel instances that do not exist in the room but are suitable for the room. It tells us what should be placed in the real-world scene and their spatial relationships when we need suggestions in the applications like augmented reality.

In this new task, the generated graph should conform to the layout habits in reality. It demands a generator that can learn and use the real-world layout experiences, which are recorded in non-empty scenes of 3D points (observations). Besides, the generated graph should meet the conditions of the current scene, such as the spatial information of existing objects. It requires the generator can extract conditional information from the current scene. Thus, this new task can be defined as a conditional generation problem in 3D space: \textbf{learning experiences from observations and using them to generate novel instances under conditions of given scenes of the point cloud}. The challenges are mainly listed below.

\begin{itemize}
  \item \textbf{Structural representation of point cloud}. Experience learning and condition extraction depend on understanding the point-cloud-based scene. It needs a structural scene representation to reflect the contextual relationships between objects. However, the point cloud, containing massive unordered points, is unstructured. Thus, a structured and order-invariant representation of the point cloud is necessary.
  \item \textbf{Real-world experience modeling}. The real-world layout experience is derived from scene information. However, scene information, mainly containing the contextual relationships between objects, is complex and high-dimensional. Thus, it is essential to find a way to model the real-world experiences that the generator can understand and use.
\end{itemize}

\textbf{The previous scene graph generation} studies jointly detect and recognize existing objects and their contextual relationships in a given scene \cite{RN212-Vision-Enhanced, RN211-Divide-and-Conquer, XuZCF17,YangLLBP18,LuKBL16,JohnsonGF18, LiJ19, SongCWJ22, RN210-Scene, RN208-Semantically}. \cite{WaldDNT20,ZhangY0021} proposed networks to represent a point-cloud-based scene as a graph with labeled nodes and semantically meaningful edges, which greatly inspired us in the structural representation of point clouds. However, these works focus on analyzing patterns of 3D scenes, lacking the ability to learn experience from scene layout.

\textbf{Instance-incremental graph generation}. Even though the proposed task of this paper has never been explored; flow-based learning \cite{PapamakariosMP17}, variational auto-encoders (VAEs) \cite{KingmaW13}, energy-based networks \cite{workhop/abs-2102-00546}, and the recursive neural network (RNN) \cite{SutskeverMH11} have led to impressive results in instance-incremental graph generation in other fields. These emerging approaches learn to model the distribution of different datasets and promote applications, such as molecular design \cite{ShiXZZZT20} and social network prediction \cite{guo2020gener}. These methods work on structured data and can not be used in instance-incremental scene graph generation from the unstructured point clouds. Recently, the image-based graph generation task has been explored \cite{usgg21, VarScene22, GEMS23}. However, the scene graph from an image records the spatial relationships in different regions of 2D views, and the graph may be discontinuous. Differently, the point cloud records all the objects in a 3D environment, and the corresponding scene graph is continuous and describes the spatial relationships in the whole scene.

In summary, existing graph generation methods are not effective for this novel task. It motivates us to explore  the following methods to tackle the above challenges. (1) \textbf{Graph representation of the point cloud}. We extract the point cloud's semantic and contextual features, designing a trainable graph to represent them. (2) \textbf{Invertible mapping between real-world scenes and simple data distributions}. Experience learning and use are inverse processes. Thus, we design an invertible mapping between the graphical scenes of real-world point clouds and the Gaussians. In this way, our generative model can learn real-world experiences through the distribution mapping of observation data in the training process, generating novel graphs for a given scene in reality through Gaussian sampling in inference.

Specifically, we propose a 3D autoregressive framework based on normalizing flows (3D-ANF) to generate novel graphs from point clouds effectively. We propose a module to learn the graph representation of the point cloud, converting objects into labeled nodes and spatial relationships into labeled edges. Besides, a normalizing flow defines the invertible mapping between graphical scenes and Gaussians. To realize the distribution mapping, a Graph Convolutional Network (GCN) effectively learns the embeddings of the subgraph (containing previously generated graphic elements); Multi-Layer Perceptrons (MLPs) map the embeddings to the mean and standard deviation of Gaussian. Based on this, we formulate the graph generation as an autoregressive process: Every node and edge generation decision is made based on the conditions extracted from previously generated elements.

Our main contributions include:

\begin{itemize}
  \item We investigate a new instance-incremental scene graph generation task from the point cloud. This task predicts scene graphs representing a given room's reasonable layouts, which help improve the realism of object arrangements in the real-world scene.
  \item We propose the first framework to realize this new task. The normalizing flows cooperate with graph representation learning to model the data distributions of the complex observations, generating graphs conforming to the layout habits in reality.
  \item We propose a new graphical dataset based on Paris-Lille-3D (Graphical-Paris-Lille-3D, GPL3D). It records the spatial relationships of objects in outdoor scenes and can be widely used in graph generation studies.
\end{itemize}

\section{Related Works}

\subsection{Graph-based Representation of Scenes}

The scene graph was first proposed in 2D computer vision by \cite{JohnsonKSLSBL15}. This work used a scene graph to represent the semantic label of objects and their relationships for image retrieval. Later, there are lines of works promoting the studies of image-based scene graphs \cite{ZellersYTC18,LiJ19,YangLLBP18,RN177, RN212-Vision-Enhanced, RN211-Divide-and-Conquer, RN210-Scene, RN208-Semantically}. These works proposed different approaches to deal with the scene graph representation problem, such as bidirectional-LSTM-based MotifNet \cite{ZellersYTC18}, a variant of GCN \cite{YangLLBP18}, a variant of Transformer \cite{RN212-Vision-Enhanced, RN211-Divide-and-Conquer}, and a combination of RNN and attention mechanism \cite{RN210-Scene}. Most of these methods relied on an object detector to extract node and edge features to compute 2D scene graphs. However, only a few works explored the representation of scene graphs in 3D point clouds due to the lack of datasets. 3RScan \cite{WaldANTN19} dataset promoted the studies of 3D scene graphs and was regarded as a new benchmark for 3D scene graph learning \cite{WaldDNT20,ZhangY0021}. \cite{WaldDNT20} proposed a novel method based on PointNet and GCN to regress a scene graph from the point-cloud-based scene. \cite{ZhangY0021} put forward an improved Edge-oriented Graph Convolutional Network (EdgeGCN) to exploit multi-dimensional edge features for explicit relationship modeling, and experiments showed promising results. All these works focused on learning the scene graph from 2D/3D data through recognizing objects and their relationships. Although different from the ultimate goal of our task, the graph representation methods inspire us a lot.

\subsection{Instance-incremental Graph Generation}

\begin{figure*}[t]
    \centering
    \includegraphics[width=1\textwidth]{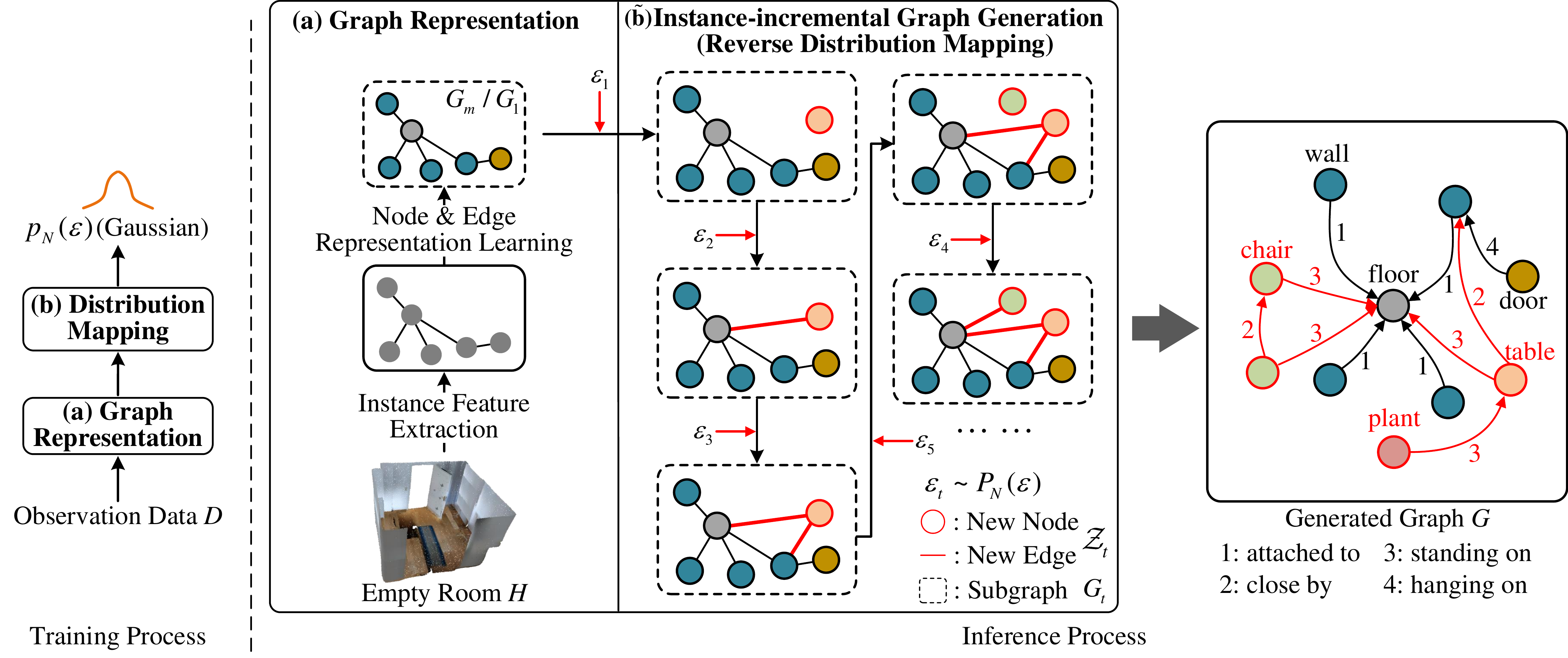}
    \caption{The overall 3D-ANF learns from observation data \emph{D} and generates novel graphs \emph{G} from empty room \emph{H}. Graphical scene ${G_m}$ is generated through instance feature extraction and representation learning. Normalizing-flow-based distribution mapping module transforms the graphical scene into Gaussian ${P_N}(\varepsilon )$ for experience modeling. In the \emph{t}-th iteration, the instance-incremental graph generation module uses experiences through sampling ${\varepsilon _t}$ from Gaussian; it creates a new node/ edge ${\mathcal{Z}_t}$ under the condition of the current subgraph ${G_t}$, where ${G_t} = {G_m},{\rm{ }}t = 1$ and ${G_t} = {G_m} \cup {{\cal Z}_{1:t - 1}},{\rm{ }}t \in \{ 2, \ldots ,T\} $; \emph{T} represents the total number of iterations. Finally, a graph describing the object layout is generated. }
    \label{Fig.2}
\end{figure*}

Similar tasks of graph generation from images have been explored before. \cite{usgg21} proposed a new task of generating unconditional scene graphs with a seed object as the input. The autoregressive method \cite{usgg21} and the VAE-based method \cite{VarScene22} learned the inner distribution of the observed images' scene graphs and generated novel scene graphs by distribution sampling. \cite{GEMS23} proposed an improved task named `Scene Graph Expansion.' This task enriched an input seed graph by adding new nodes and the corresponding relationships. The authors also proposed an autoregressive method with an improved sequencing strategy to tackle this task.

Things make a difference between ours and the above image-based task. Our task outputs \textbf{a continuous graph} to describe the continuous spatial relationships in a global 3D scene, and all the graphic elements are within a whole graph. However, the above image-based tasks describe the spatial relationships in different regions of 2D views---the output may be \textbf{a discontinuous graph} consisting of isolated nodes/triples or subgraphs.

Numerous works applied deep models of instance-incremental graph generation in other fields, especially in molecular design \cite{MaCX18, workhop/abs-2102-00546, YouLYPL18} and social network prediction \cite{GroverZE19, TranSSG22}. \cite{MaCX18, GroverZE19} respectively proposed VAE-based frameworks to generate semantically valid graphs used in molecule generation and link prediction of social networks. \cite{workhop/abs-2102-00546} designed the first energy-based model GraphEBM for generating molecular graphs. It applied Langevin dynamics to train a graph's energy function by maximizing the likelihood and generating samples with low energies. The state-of-the-art models are built based on autoregressive approaches. \cite{YouLYPL18} formulated the graph generation problem as a sequential decision process and added new nodes and edges based on current subgraph structures. GraphRNN \cite{YouYRHL1811} tackled this sequential decision problem using RNN, and the improved MolecularRNN \cite{abs-1905-13372} added the judgment of semantic labels of edges.

Recently, autoregressive generative models with normalizing flows have achieved good experimental results in many tasks \cite{ShiXZZZT20,LuoYJ21}. This flow-based approach maps the graph data to a latent base distribution (e.g., Gaussian). The invertible transformation gives the model a high capacity to model high-dimensional data. However, these methods do not consider the constraints of real-world entities, leaving gaps to be used in graph generation based on real-world 3D scenes. Our study fills the gap and introduces the flow-based model to instance-incremental graph generation in 3D space.

\section{Problem Statement}

The instance-incremental graph generation from real-world point clouds can be formulated as follows: Given a point-cloud-based scene $H \in {\mathbb{R}^{n \times c}}$ including \emph{n} points with \emph{c} channels, outputting a layout graph \emph{G} suitable for \emph{H} and conforming to layout habits in reality. To achieve this goal, we explore a way to model real-world layout experiences from observed non-empty room scans. This modeling process can be denoted as $D \to P$, where \emph{D} represents the observations and \emph{P} is the modeled distribution. Based on this, a novel graph is generated using real-world experiences and considering the conditions of the current scene, which can be represented as $H,P \to G$.

\section{Approach}

\subsection{Overview}

We employ a two-stage framework to tackle this generation problem, as presented in Fig. \ref{Fig.2}. In stage (a), the \textbf{\emph{graph representation}} module, learned from observation data, converts the rooms of 3D points into graphs. In stages (b) and (${\rm{\tilde b}}$), the \textbf{\emph{distribution mapping}} module captures observations' inner object layout experiences; The invertible process, \textbf{\emph{graph generation}}, leverages the modeled experiences to create graphic elements under the conditions of the current subgraph. All the details will be discussed below.

\subsection{Graph Representation of Point Cloud}

\begin{figure}[htbp]
    \centering
    \includegraphics[width=0.46\textwidth]{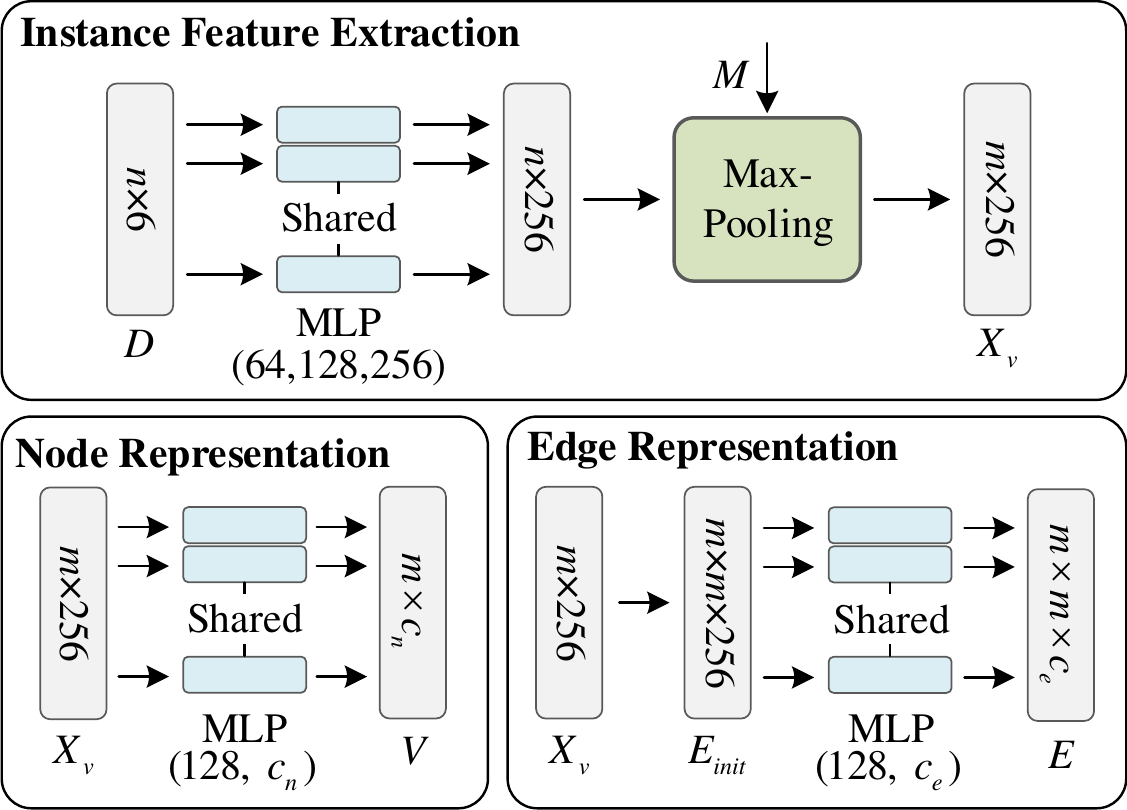}
    \caption{\textbf{Graph representation learning}. A point cloud \emph{D} is considered the input, and the graphical scene ${G_m} = (V,E)$ is considered the output. BatchNorm is used for these MLPs with ReLU; Softmax works after the final layer of MLP in node and edge representation learning. \emph{M} is given in the dataset.}
    \label{Fig.3}
\end{figure}

The structural representation of the point cloud is the basis of the learning experiences and extracting conditions from scenes. To achieve the structural representation, we explore a way to convert the unordered points to a semantically meaningful graph denoting instances with relationships. As illustrated in Fig. \ref{Fig.3}, we propose a module to extract the features of 3D instances (\emph{wall}, \emph{floor}, etc.) and learn the node/edge representation.

Specifically, parameter-sharing multi-layer perceptron (MLP) captures the point-wise features of the input point cloud \emph{D}. Then, the max-pooling operates on the unordered encoded point cloud, along with the class-agnostic point-to-instance indicator $M \in {\{ 1, \ldots ,m\} ^n}$ \cite{RN78,ZhangY0021}, resulting in the order-invariant encoding of the included \emph{m} instances. ${X_v} \in {\mathbb{R}^{m \times 256}}$ denotes the instance-wise feature with 256 channels.

${X_v}$ is further propagated to facilitate the modeling of the representation of \emph{m} nodes and ${m^2}$ one-to-one edges in ${G_m} = (V,E)$, as $V \in {\mathbb{R}^{m \times {c_n}}}$ and $E \in {\mathbb{R}^{m \times m \times {c_e}}}$. ${c_n}$ and ${c_e}$ equal to the number of object classes and relationship classes, respectively. In node representation learning, MLP performs on ${X_v}$ for node feature evolution, converting the channel number from 256 to the final ${c_n}$. In edge representation learning, we introduce an edge feature matrix ${E_{init}} \in {\mathbb{R}^{m \times m \times 256}}$ to establish the initial relationships between instances, where ${E_{init}}[i,j,:] = {X_v}[i,:] - {X_v}[j,:]$ and $i,j \in \{ 0, \ldots ,m - 1\} $. MLP performs on ${E_{init}}$ for edge feature evolution, resulting in the final edge representation.

\subsection{Invertible Distribution Mapping}

One of the core problems in this novel task is how to generate graphic elements (nodes/edges) ${\mathcal{Z}_t}$ under the conditions of the current subgraph ${G_t}$. This autoregressive conditional probability can be denoted as $P({z_t}|{G_t})$, where ${z_t}$ is the representation of ${\mathcal{Z}_t}$ with the one-hot label, and \emph{t} refers to the \emph{t}-th iteration. However, the graphic elements are discrete and follow complex data distributions. It poses challenges to calculate $P({z_t}|{G_t})$.

\textbf{Normalizing flows} define an invertible mapping between samples with complex distributions and samples conforming to base distributions \cite{PapamakariosMP17}. It can map a complex graphic element ${z_t}$ into a sample ${\varepsilon _t}$ of Gaussian, achieving the data distribution modeling of graphic elements in the training stage. Besides, it can map a sample ${\varepsilon _t}$ of the Gaussian back to a graphic element ${z_t}$ in the generation stage.

We introduce it into our task and denote $P({z_t}|{G_t})$ as a Gaussian process:

\begin{equation}
  P\left( {{z_t}\mid {G_t}} \right) = {\cal N}\left( {{\mu _t},{{\left( {{\sigma _t}} \right)}^2}} \right)
  \label{eq1}
\end{equation}

Where ${\mu _t}$ and ${\sigma _t}$ are the Gaussian parameters decided by the current subgraph ${G_t}$ (further discussed in the \textbf{Condition Evaluation} paragraph). Through an affine transformation \cite{ShiXZZZT20}, ${z_t}$ can be converted to ${\varepsilon _t}$ conforming to the standard Gaussian ${\cal N}\left( {0,1} \right)$.

\begin{equation}
  \left\{ {\begin{array}{*{20}{c}}
{{f^{ - 1}}\left( {{z_t}} \right) = {\varepsilon _t} = \frac{{{z_t} - {\mu _t}}}{{{\sigma _t}}}}\\
{f\left( {{\varepsilon _t}} \right) = {z_t} = {\mu _t} + {\sigma _t} \cdot {\varepsilon _t}}
\end{array}} \right.
  \label{eq2}
\end{equation}

\begin{figure}[htbp]
    \centering
    \includegraphics[width=0.475\textwidth]{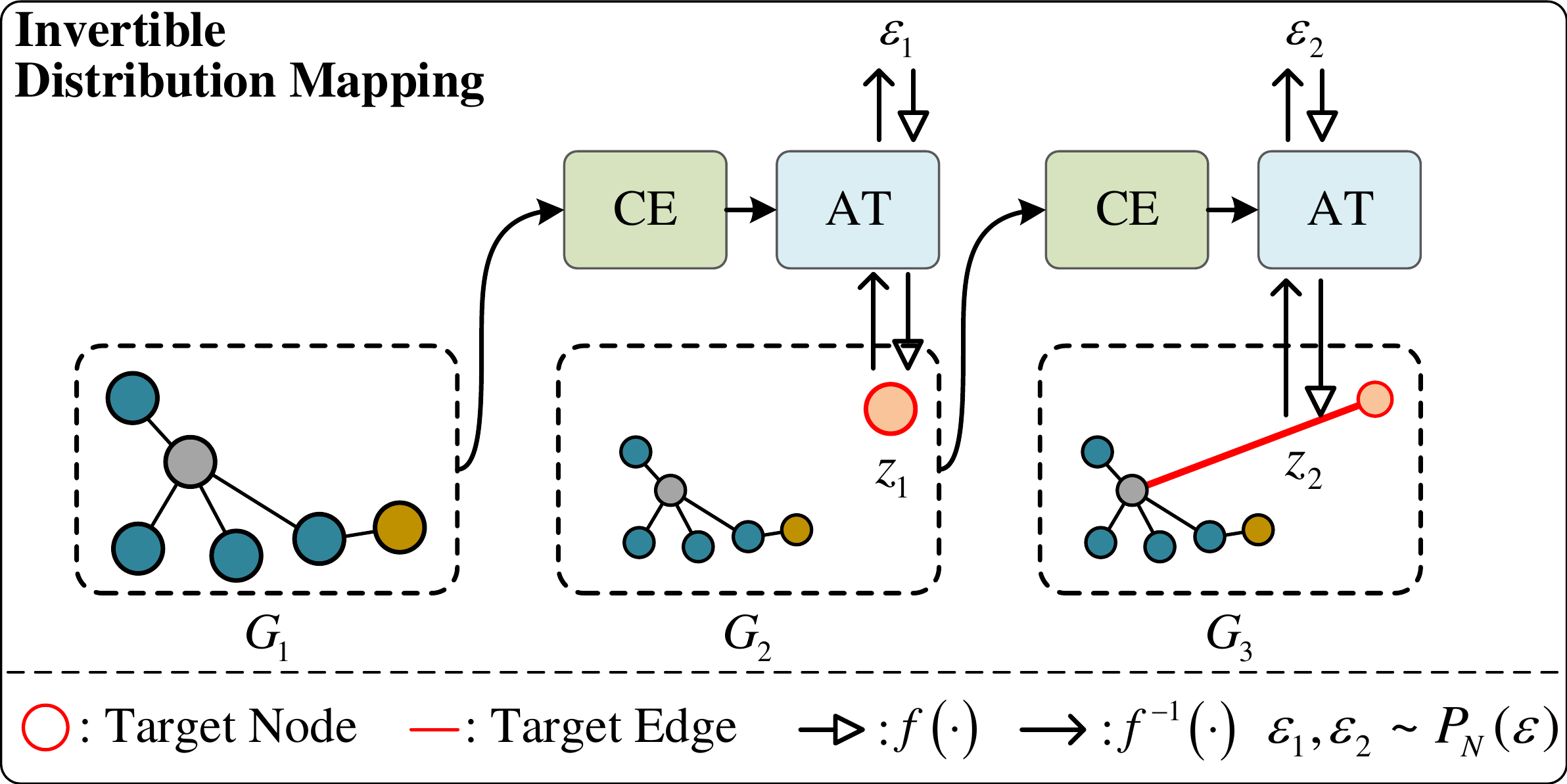}
    \caption{\textbf{Invertible distribution mapping via normalizing flows}. \emph{Affine Transformation} (AT) and \emph{Conditional Evaluation} (CE) are the cores of this module. The former bridges between the target node/edge ${z_t}$ and value ${\varepsilon _t}$ following the standard Gaussian ${P_N}\left( \varepsilon  \right)$; the latter extracts conditions from the current subgraph ${G_t}$.}
    \label{Fig.4}
\end{figure}

We explain how to make equation (\ref{eq1}) work and the details of the invertible distribution mapping in Fig. \ref{Fig.4}. As normalizing flows work on continuous data, thus we first convert discrete ${z_t}$ into continuous data, where $\alpha$ is the weight coefficient of $\gamma \sim U[0,1)$ \cite{WOS000461852004074}.

\begin{equation}
{z_t} = {z_t} + \alpha  \cdot \gamma
\label{eq3}
\end{equation}

\emph{In the training phase}, condition evaluation functions work on $G_t$, resulting in Gaussian parameters ($\mu _t$, $\sigma _t$). Through the affine function $f^{-1}( \cdot )$, $z_t$ is mapped to $\varepsilon _t$. The equation (\ref{eq1}) is workable if all the $\varepsilon _t$ accord with the standard Gaussian. The normalizing flows method calculates the distribution density behind equation (\ref{eq1}) as \cite{ShiXZZZT20}:

\begin{equation}
{p_Z}\left( {{z_t}} \right) = {p_N}\left( {{f^{ - 1}}\left( {{z_t}} \right)} \right)\left| {\det \frac{{\partial {f^{ - 1}}\left( {{z_t}} \right)}}{{\partial {z_t}}}} \right|
\label{eq4}
\end{equation}

Where ${p_Z}\left(  \cdot  \right)$ is the exact density of the graphic elements, ${p_N}\left(  \cdot  \right)$ is the density of standard Gaussian, $\det \frac{{\partial {f^{ - 1}}\left( {{z_t}} \right)}}{{\partial {z_t}}}$ is the Jacobian determinant. The Gaussian transformation is achieved by minimizing the negative log-likelihoods of equation (\ref{eq4}).

\emph{In the t-th iteration of generation (inference time)}, $\varepsilon _t$ is sampled from ${P_N}(\varepsilon )$. A novel node or edge $z_t$ is generated through ${f}\left( {{\varepsilon _t}} \right) = {z_t} = {\mu _t} + {\sigma _t} \cdot {\varepsilon _t}$, where ($\mu _t$, $\sigma _t$) are calculated according to the current subgraph.

\begin{figure}[htbp]
    \centering
    \includegraphics[width=0.475\textwidth]{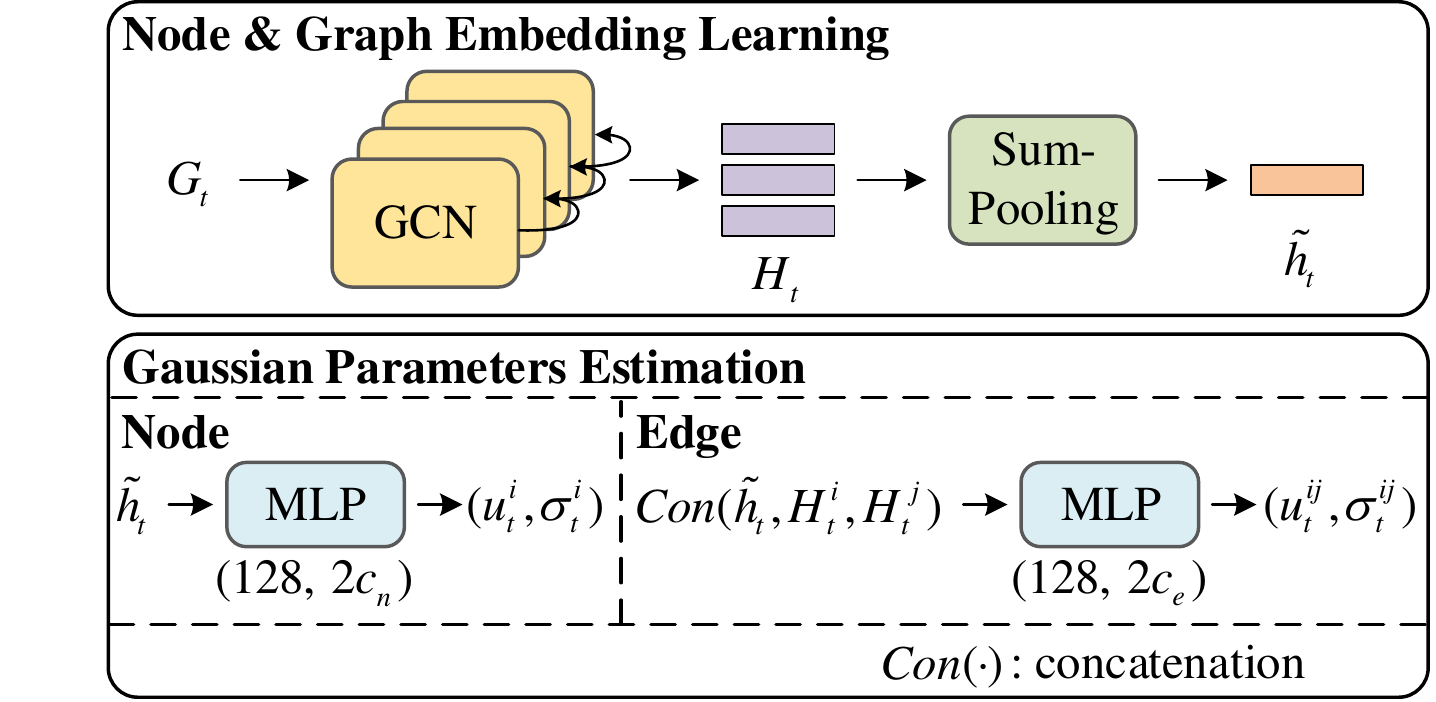}
    \caption{\textbf{Conditional evaluation}. This submodule considers the subgraph ${G_t}$ as input. It first learns the node and graph embeddings and then calculates the Gaussian parameters.}
    \label{Fig.5}
\end{figure}

\textbf{Conditional Evaluation} aims at calculating Gaussian parameters ($\mu _t$, $\sigma_t$) from the current subgraph $G_t$, which should fully consider the characteristics of the subgraph and the relationships between graphic elements. To achieve this goal, as presented in Fig. \ref{Fig.5}, a 4-layer GCN learns the node embeddings ${H_t} \in {\mathbb{R}^{{m_t} \times 128}}$, cooperating with sum-pooing to obtain the subgraph embedding ${\tilde h_t} \in {\mathbb{R}^{128}}$, where ${m_t}$ indicates the number of nodes in ${G_t}$. Based on this, a branch projects the subgraph embedding to Gaussian parameters ($u_t^i$, $\sigma _t^i$) in the generation of node \emph{i}. Another branch outputs the Gaussian parameters ($u_t^{ij}$, $\sigma _t^{ij}$) in the generation of the edge between nodes \emph{i} and \emph{j}, considering both the embedding of the subgraph ${\tilde h_t}$ and the embeddings of adjacent nodes ($H_t^i$, $H_t^j$) \cite{LuoYJ21}.

\subsection{Loss Functions}
\textbf{Graph representation loss}. Two multi-class cross-entropy losses ${L_n}$ and ${L_e}$ guide the representation learning with label semantics. ${L_n}$ is used for node representation learning while ${L_e}$ for the edge. Note that we add an `\emph{empty}'  class in the representation learning for the edge $E \in {\mathbb{R}^{m \times m \times {c_e}}}$, indicating no adjacency between two nodes. It was discovered that the \emph{empty} edges generally account for the majority of edges in the training dataset, bringing a class-imbalanced problem. Inspired by \cite{ZhangY0021}, the adjacency matrix \emph{R} is given in the dataset. It helps to ignore the \emph{empty} edge in the learning process to solve the problem.

\textbf{Distribution mapping loss}. Equation (\ref{eq4}) formulates the exact probability density of a graphic element based on an invertible distribution mapping. To train the normalizing-flow-based model for the distribution mapping, we compute the log-likelihoods of all the \emph{N} graphic elements and update model parameters by minimizing the negative log-likelihoods \cite{LuoYJ21,RN207-Topology-Aware}:

\begin{equation}
\begin{aligned}
{L_m} & =  - \sum\limits_{t = 1}^N {\log {p_Z}\left( {{z_t}} \right)} \\
& = - \sum\limits_{t = 1}^N {[\log {p_N}\left( {{f^{ - 1}}\left( {{z_t}} \right)} \right) + \log \left| {\det \frac{{\partial {f^{ - 1}}\left( {{z_t}} \right)}}{{\partial {z_t}}}} \right|]}
\end{aligned}
\label{eq5}
\end{equation}

Thus, the joint loss function to train the overall model is defined as:

\begin{equation}
L = {L_n} + {L_e} + {L_m}
\label{eq6}
\end{equation}

\subsection{Details of Instance-incremental Graph Generation}

\begin{figure}[htbp]
    \centering
    \includegraphics[width=0.475\textwidth]{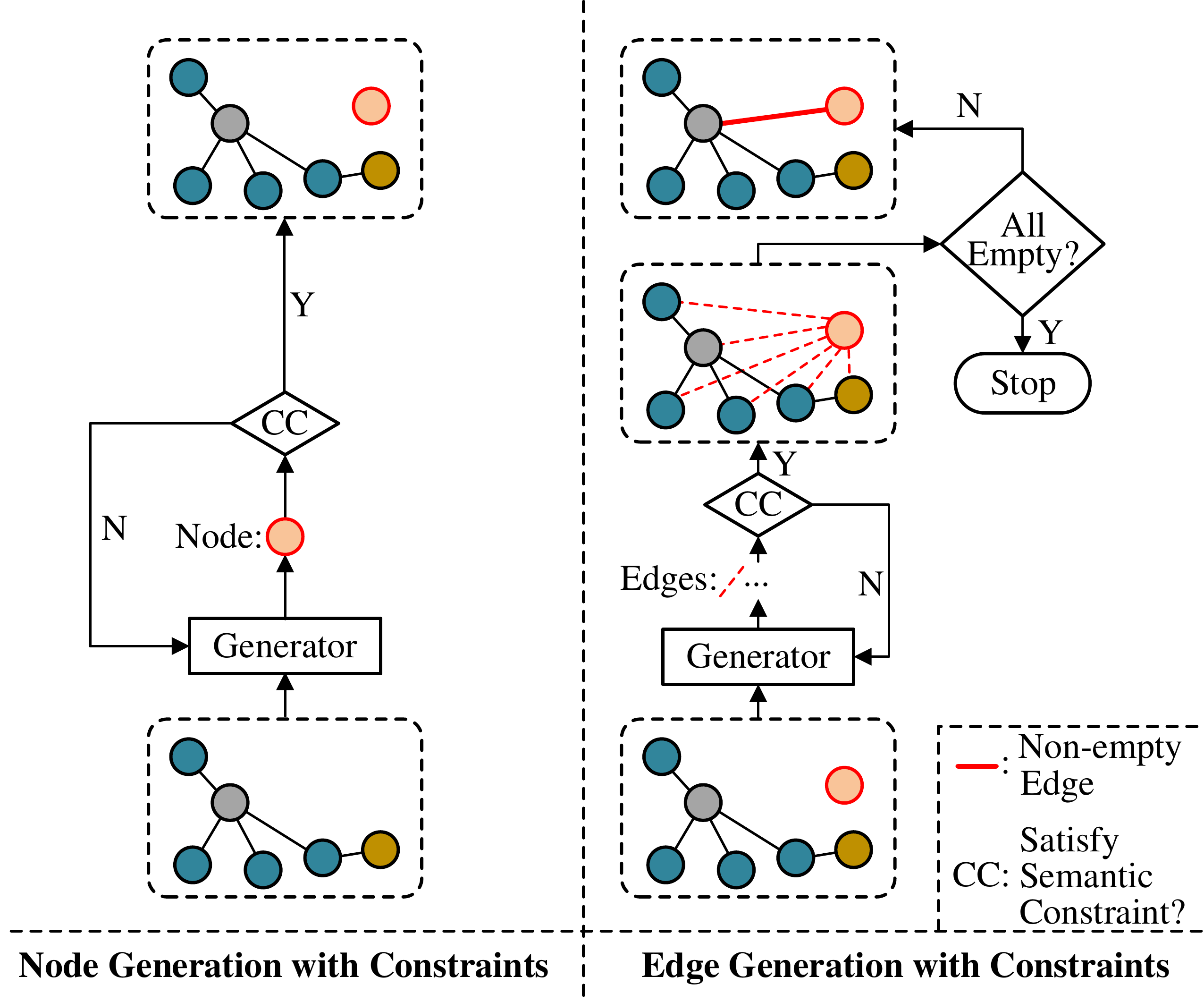}
    \caption{\textbf{Node/edge generations with semantic constraints}. Our method starts with node generation, then establishes the edge between this new node and the existing ones. These above processes repeat. It stops if all the predicted edges are \emph{empty}.}
    \label{Fig.6}
\end{figure}

This section discusses (1) \emph{How to make the graph more plausible}? (2) \emph{When to generate a node \& when to generate an edge}? (3) \emph{When does the graph generation stop}?

To make the graphs more plausible, we consider adding constraints to the generation. 1) Space constraints. The total volume of generated entities needs to be within a reasonable range so that the space does not appear too crowded or sparse. 2) Semantic constraints. The semantic labels of generated entities should be reasonable. For example, a \emph{floor} should not be generated in an enclosed room, as well as a \emph{bed} in the \emph{kitchen}. 3) Anti-overlapping operations. If we can give the suggested object poses, it can help prevent the object from overlapping in the downstream applications. Section $\uppercase\expandafter{\romannumeral5}$-B shows the implementation details.

Taking the node/edge generations with semantic constraints as examples (Fig. \ref{Fig.6}). Our method starts by generating a new node and determines whether it satisfies the semantic constraint. If satisfied, the node is added to the graph; otherwise, it is abandoned, and a new node is generated. Then, our method establishes the edges between the new node and existing nodes. If all the edges are \emph{empty}, the generation is stopped; otherwise, we keep the \emph{non-empty} edges and repeat the above processes. Experiments will demonstrate the generation results with constraints. Supplementary material-1 describes more details through pseudocode.

\section{Experiment}

To evaluate the effectiveness of our proposed method on this new graph generation task, we run experiments on the benchmark dataset and our newly proposed dataset. Experiments are designed to answer the following research questions: (1) \emph{Can the proposed 3D-ANF achieve this new generation task?} (2) \emph{How does 3D-ANF perform compared to other graph generation methods?} (3) \emph{Do several important designs of 3D-ANF affect the generation results?} (4) \emph{How does adding constraints impact the graph generation?} (5) \emph{Can 3D-ANF work well on outdoor scans?}

Sections \emph{A} to \emph{C} introduce the dataset, implementation details, and evaluation metrics of our experiment. Sections \emph{D} and \emph{E} answer questions (1) and (2) through comparison and qualitative experiments. Section \emph{F} verifies the effectiveness of several designs through an ablation study, which answers question (3). Section \emph{G} answers question (4) by quantifying the experimental results of generation with constraints. Section \emph{H} answers question (5) by conducting experiments on our newly proposed outdoor graphical dataset GPL3D.

\subsection{Dataset}

There are not many datasets for the scene graph generation tasks from point clouds. Only 3DSSG-O27R16 \cite{ZhangY0021}, a dataset of indoor scenes, can be used as the benchmark. It combines 3DSSG \cite{WaldDNT20} and 3RScan \cite{WaldANTN19}, containing 1318 non-empty rooms in point clouds and the corresponding scene graph. Each point cloud provides XYZ coordinates, RGB values, etc. The semantic annotations of instances, including 27 object classes (e.g., \emph{lamp} and \emph{ceiling}) and 16 non-empty relationship classes (e.g., \emph{hanging on} and \emph{attached to}), are recorded in the scene graph. We remove 64 scenes from the dataset. These scenes include more than 50 instances and cost excessive computing resources in the training process. We randomly choose about 255 scenes as the test data. We remove the semantic annotations and the entities inside the room from each test data, only retaining an empty room with XYZ coordinates and RGB values for the instance-incremental graph generation. Besides, to verify the effectiveness of this novel task and the performance of methods on outdoor scans, we have made a new dataset GPL3D, and the details are in Section $\uppercase\expandafter{\romannumeral5}$-H.

\subsection{Implementation Details}

The proposed network is end-to-end trained using back-propagation and Adam optimizer with an initial learning rate of 0.001. The experiments are conducted by Pytorch and Dive into Graphs (DIG) \cite{liu2021dig} on a single GeForce GTX 1080 Ti and an Intel(R) Core(TM) i7-7800X CPU; The training batch size is set to 32. The weighting value $\alpha$ in Equation (\ref{eq3}) is set to 0.9; a 4-layer GCN is used to learn the graph and node embeddings. All these details will be discussed in the ablation studies. The graph generation starts from a random value sampled from the Gaussian, bringing randomness to the results. Each test data generates five graphs for evaluation to reduce the impact of randomness on the results.

Some data processing and statistical analysis are performed to support the generation with constraints. 1) \textbf{Space constraints}. We take the following measures to introduce space constraints into the generation. Firstly, we calculate the volume of each room in the test data. Next, we summarize the volume data of each entity in the dataset and model the per-category volume distribution as Gaussian. In graph generation, the volume of each generated entity is calculated through sampling from the corresponding Gaussian. The generation stops when the total volume of the generated entities is beyond a certain proportion of the room's volume. 2) \textbf{Semantic constraints} aim to restrict the categories of objects generated in different kinds of rooms. On the one hand, we manually label the room functions in the test data, such as the \emph{living room} and \emph{office}. On the other hand, we establish a dictionary to record the relationships between the room functions and the object categories in the real world. If the category of a generated instance breaks the rules recorded in the dictionary, it will be regarded as invalid and needs to be regenerated. 3) \textbf{Anti-overlapping operations}. We consider the corresponding 3D object as a bounding box and randomly give the pose information according to the spatial relationships in the graph. If another object has occupied the place, reproduce the object pose; If a suitable object location cannot be found after several iterations, the node is abandoned, and a node with relational semantics is newly generated.

\begin{table*}[htbp]
\centering
\caption{Quantitative results on 3DSSG-O27R16. $\uparrow$: Larger is better; $\downarrow$: Lower is better.}
\label{table1}
\begin{tabular}{c|cccccc}
\hline
\multirow{2}{*}{Method} & \multirow{2}{*}{\begin{tabular}[c]{@{}c@{}}Node\\ Validity (\%) $\uparrow$\end{tabular}} & \multirow{2}{*}{\begin{tabular}[c]{@{}c@{}}Edge\\ Validity (\%) $\uparrow$\end{tabular}} & \multicolumn{2}{c}{MMD $\downarrow$} & \multirow{2}{*}{\begin{tabular}[c]{@{}c@{}}Uniqueness\\ \% $\uparrow$\end{tabular}} &
\multirow{2}{*}{\begin{tabular}[c]{@{}c@{}}Diversity\\ \% $\uparrow$\end{tabular}}\\ \cline{4-5}
 &  &  & Degree & Cluster &  \\ \hline
GraphEBM & 79.6 & 13.4 & 0.86 & 1.47 & \textbf{100.0} & \textbf{80.5}\\
GraphRNN & \textbf{80.6} & 30.6 & 0.93 & 0.78 & 71.2 & 26.1\\
SGG-GEMS & 77.6 & 39.2 & 0.68 & 0.85 & 83.1 & 32.6\\
3D-ANF (Ours) & 78.4 & \textbf{84.7} & \textbf{0.44} & \textbf{0.08} & \textbf{100.0} & 69.3\\ \hline
\end{tabular}
\end{table*}

\subsection{Evaluation Metrics}

Our instance-incremental task outputs the graphs representing the suitable object layout schemes for a real-world scene. The layout schemes can be various. Thus, our task does not have a unique solution. A reasonable layout should accord with real-world layout habits and be consistent with human perception:

From the perspective of each generated instance, the instance should conform to the objective reality. For a completely enclosed space, the generated objects should be furniture, such as \emph{tables} and \emph{sofas}, rather than the existing entities, such as \emph{walls} and \emph{floors}. Besides, `a chair \emph{standing on} the ceiling.' is obviously wrong. Thus, we judge whether each generation conforms to the objective reality and introduce \textbf{Node/Edge Validity} to quantify the proportion of correctly generated nodes/edges in all the generated ones \cite{ShiXZZZT20}.

From the global perspective, the gap between the generated layouts and the layouts of reality should be small. \textbf{Maximum Mean Discrepancy (MMD)} indicates the similarity between two data distributions \cite{GrettonBRSS12,YouYRHL1811}. We introduce MMD based on degrees and clustering coefficients to quantify the similarity between the generated and real-world layout graphs. The real-world ones are obtained through the manual annotation of numerous non-empty rooms of reality. The smaller the MMD, the smaller the gap between the generated layouts and the layouts of reality.

\textbf{Uniqueness} and \textbf{Diversity} metrics are introduced to quantify the graph difference. Uniqueness describes the percentage of graphs different from others \cite{ShiXZZZT20}. Diversity is a stricter metric. For each test scene, we exclude the graph that is a subgraph of the other graphs, and diversity equals the remaining percentage \cite{GEMS23}. Besides, we evaluate the complexity of methods from two aspects: the \textbf{Model Parameters (Params)} and the \textbf{Floating Point Operations (FLOPs)}

\subsection{Comparison with Baselines}

\textbf{Baselines}. Existing instance-incremental graph generation methods cannot be directly used in this new task, as described above. Thus, we introduce the widely used methods in other fields, the energy-based GraphEBM \cite{workhop/abs-2102-00546} and the RNN-based GraphRNN \cite{YouYRHL1811,abs-1905-13372}, to cooperate with our point cloud graph representation as baselines. Note the VAEs-based method, such as JT-VAE \cite{JinBJ18}, also works well in the instance-incremental generation of other fields. However, this kind of method relies on prior knowledge to guide the generation, which is unfair for the comparison. Thus, we do not consider it as the baseline method. Besides, we introduce a method consisting of the graph representation network SGG${\rm_{point}}$ \cite{ZhangY0021} and the graph expansion network GEMS \cite{GEMS23} as the baseline, verifying if the image-based graph expansion method can be used in our point-cloud-based task. We call this baseline method SGG-GEMS.

\textbf{Node/Edge Validity}. Quantitative results using different methods are illustrated in TABLE \ref{table1}. On the one hand, our method significantly improves the edge validity, 45.5\% higher than the SGG-GEMS in the second place. On the other hand, there is little performance difference between these methods in terms of node validity.

The validity metrics indicate the understanding ability of real-world scenes, especially for the included semantic information. Unlike the graphs with simple semantics, such as molecules, the node/edge semantics in a point-cloud-based scene is complicated.

The 3DSSG-O27R16 dataset contains 27 node classes and 17 edge classes (including the empty one). In the real-world experience learning stage, every method learns the data distribution from every scene's node features $V \in {\mathbb{R}^{m \times {c_{n}}}}$ and edge features $E \in {\mathbb{R}^{m \times m \times {c_{e}}}}$, where ${c_n} = 27$, ${c_e} = 17$, and \emph{m} equals the max object number 50. The edge feature matrix is much more complex than the node one, containing about 30 times as many elements as the latter, and the values are more dispersed. Thus, the data distribution learning of edges is more challenging than that of nodes.

All the methods work well in node generation because the data distribution learning of nodes is relatively easy. Even though the misjudgment of the exited objects' semantics may bring mistakes to the generation of novel nodes, the node validities of all methods remain high. Things make a difference in the edge generation. GraphEBM and GraphRNN model the observations as an energy function or a sequence, respectively. These methods cannot model the complex data distributions of edges well, resulting in poor validities of generated edges. The improved sequencing strategy of SGG-GEMS improves this situation but still retains a low level. Our 3D-ANF transforms the observations into Gaussians with parameters adjusted according to conditions. It has a strong ability to model the data distributions and promotes the validities of graphic elements.

\textbf{MMD}. TABLE \ref{table1} illustrates that our method obtains the best MMD scores for degrees and clustering coefficients. It indicates that our generated graphs are closer to the real-world scene data.

The quality of generated graphs depends on the real-world experiences of object layouts learned from observation data. GraphRNN represents graphs under different orderings as sequences, and the improved SGG-GEMS introduces a novel method and external knowledge to improve the layout experience learning. However, on the one hand, these sequence-based methods cannot model the in-depth features of complex object layouts well. On the other hand, SGG-GEMS, an image-based method, often outputs a graph containing several subgraphs or isolated nodes/triples to denote the layout of a 3D room. It is inconsistent with the logic of reality, leaving an obvious gap between the generations and the observations. The energy-based GraphEBM also works not well because of the poor ability of layout experience learning. Unlike the baselines, our method learns the node and graph embeddings through GCN. It aggregates the features of neighbors through message-passing layers, which well models the layout information represented in the graph. Thus, our method performs well in terms of MMD scores.

\textbf{Uniqueness $\&$ Diversity}. TABLE \ref{table1} illustrates that our method and GraphEBM achieve 100\% scores on graph uniqueness. However, the graphs generated by GraphEBM exhibit greater diversity than ours. GraphRNN and SGG-GEMS achieve the lowest score of graph uniqueness and diversity.

The real-world layout experiences constrain the objects placed in the scene and their spatial relationships. The poor edge validity and MMD metrics indicate that GraphEBM performs poorly in learning the layout experiences, which results in weak constraints from reality in graph generation. Thus, the graphs may be unreasonable but diverse. Differently, our method achieves a good balance between graph rationality and diversity. The graphs accord with the real-world layout habits and also retain an acceptable score on graph diversity.

In summary, our proposed 3D-ANF achieves this novel instance-incremental graph generation and outputs reliable novel graphs; besides, our method outperforms other baselines in most evaluation metrics.

\textbf{Method Complexity}. TABLE \ref{table1.5} shows that our 3D-ANF has the second least number of model parameters and the fewest FLOPs. The sequence-based GraphRNN and SGG-GEMS suffer from relatively dense parameters and computations. They focus on modeling the relationships between elements in the sequence but cannot extract the layout habits behind the relationships well. The energy-based GraphEBM has the least number of model parameters. However, it needs a high computing cost to evaluate the energy value of the graph to be generated. By comparison, our efficient method achieves the best graph generation performance at a low cost.

\begin{table}[htbp]
\centering
\caption{Comparison of method complexity.}
\label{table1.5}
\begin{tabular}{c|cc}
\hline
Method & Params (M) $\downarrow$ & FLOPs/Scene (G) $\downarrow$ \\ \hline
GraphEBM & \textbf{0.31} & 4.11 \\
GraphRNN & 2.11 & 2.12 \\
SGG-GEMS & 3.07 & 2.70 \\
3D-ANF (Ours) & 1.27 & \textbf{1.60} \\ \hline
\end{tabular}
\end{table}

\subsection{Qualitative Analysis}

\begin{figure*}[htbp]
    \centering
    \includegraphics[width=0.9\textwidth]{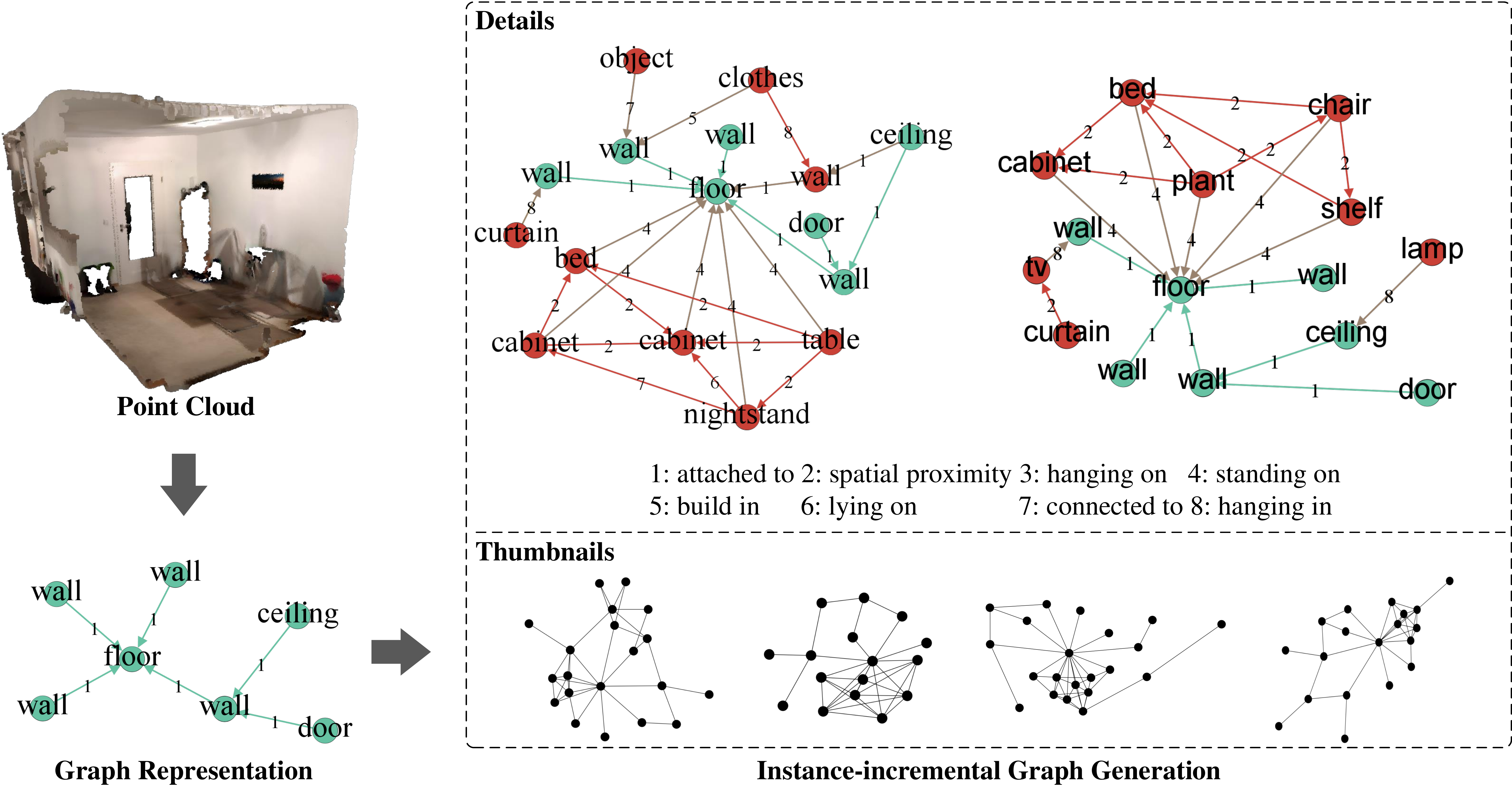}
    \caption{The visualization of the instance-incremental graph generation from a point cloud. Our method first predicts the graph representation of an empty room in the point cloud, then generates proposed furniture (nodes) and spatial relationships (edges). The green nodes and edges in the detailed graph indicate the instances and the spatial relationships already existing in the scene, while other colors indicate the newly generated elements. Thumbnails show several graphs generated from the same scene.}
    \label{Fig.7}
\end{figure*}

Fig. \ref{Fig.7} shows the visualization of the graph generation from an empty room in the point cloud. Our method extracts conditions from the current graphical scene, autoregressively generating novel nodes (e.g., \emph{plant}) and edges (e.g., `a plant \emph{standing on} the floor.') with real-world layout experiences. As a result, a graph describing the proposed object layout is created.

Most generations conform to the layout habits in reality. However, some invalid generations have occurred, e.g., `\emph{clothes build in the wall}.' The reasons are as follows: each instance is generated considering both the real-world layout experiences and the conditions of the current scene. However, on the one hand, the layout experiences may not be fully understood and modeled; on the other hand, experiences come from observations, which have a distribution different from the data used for inference. The layout experiences may be unsuitable for some scenes, thus causing invalid generations. Fig. \ref{Fig.7} also shows several generated graphs of the same scene. All these graphs have unique structures and semantics, providing a variety of schemes for object layouts.

\subsection{Ablation Study}

\begin{table*}[htbp]
\centering
\caption{Ablation study of several designs.}
\label{table4}
\begin{tabular}{c|c|c|cccccc}
\hline
\multirow{2}{*}{\begin{tabular}[c]{@{}c@{}} $\alpha$ in\\ Equation (2)\end{tabular}} & \multirow{2}{*}{\begin{tabular}[c]{@{}c@{}}GCN\\ Layers\end{tabular}} & \multirow{2}{*}{\begin{tabular}[c]{@{}c@{}}Cross-entropy\\ Loss\end{tabular}} & \multirow{2}{*}{\begin{tabular}[c]{@{}c@{}}Node\\ Validity (\%) $\uparrow$\end{tabular}} & \multirow{2}{*}{\begin{tabular}[c]{@{}c@{}}Edge\\ Validity (\%) $\uparrow$\end{tabular}} & \multicolumn{2}{c}{MMD $\downarrow$} & \multirow{2}{*}{\begin{tabular}[c]{@{}c@{}}Uniqueness\\ \% $\uparrow$\end{tabular}} & \multirow{2}{*}{\begin{tabular}[c]{@{}c@{}}Diversity\\ \% $\uparrow$\end{tabular}} \\ \cline{6-7}
 &  &  &  &  & Degree & Cluster &  \\ \hline
0.6 & \multirow{5}{*}{4} & \multirow{7}{*}{w/} & \textbf{78.5} & \textbf{93.4} & 0.63 & 0.22 & 99.6 & 51.9\\
0.8 &  &  & 76.3 & 87.2 & 0.65 & 0.23 & 99.9 & 53.2\\
0.9 &  &  & 78.4 & 84.7 & 0.44 & \textbf{0.08} & \textbf{100.0} & 69.3\\
1.0 &  &  & 75.0 & 55.8 & \textbf{0.34} & 0.10 & \textbf{100.0} & \textbf{72.1}\\
1.2 &  &  & 75.9 & 59.6 & 0.70 & 0.25 & \textbf{100.0} & 59.3 \\ \cline{1-2}
\multirow{3}{*}{0.9} & 3 &  & 77.6 & 84.7 & 0.55 & 0.17 & 99.9 & 64.5\\
 & 5 &  & 76.8 & 59.9 & 0.70 & 0.25 & \textbf{100.0} & 51.0\\ \cline{3-3}
 & 4 & w/o & 78.1 & 65.7 & 0.61 & 0.19 & \textbf{100.0} & 60.5\\ \hline
\end{tabular}
\end{table*}

\begin{table*}[htbp]
\centering
\caption{Quantitative results of generation with constraints.}
\label{table5}
\centering
\begin{tabular}{cc|c|c|cccccc}
\hline
\multicolumn{2}{c|}{\multirow{2}{*}{\begin{tabular}[c]{@{}c@{}}Space\\ Constraint\end{tabular}}} & \multirow{3}{*}{\begin{tabular}[c]{@{}c@{}}Type of\\ Semantic\\ Constraint\end{tabular}} & \multirow{3}{*}{\begin{tabular}[c]{@{}c@{}}Anti-\\ overlapping\end{tabular}} & \multirow{3}{*}{\begin{tabular}[c]{@{}c@{}}Node\\ Validity (\%) $\uparrow$\end{tabular}} & \multirow{3}{*}{\begin{tabular}[c]{@{}c@{}}Edge\\ Validity (\%) $\uparrow$\end{tabular}} & \multicolumn{2}{c}{\multirow{2}{*}{MMD $\downarrow$}} & \multirow{3}{*}{\begin{tabular}[c]{@{}c@{}}Uniqueness\\ \% $\uparrow$\end{tabular}} & \multirow{3}{*}{\begin{tabular}[c]{@{}c@{}}Diversity\\ \% $\uparrow$\end{tabular}} \\
\multicolumn{2}{c|}{} &  &  &  &  & \multicolumn{2}{c}{} &  &  \\ \cline{1-2} \cline{7-8}
$\lambda$ & $\beta$ &  &  &  &  & Degree & Cluster &  &  \\ \hline
\multirow{3}{*}{0.8} & \multirow{3}{*}{1.2} & 1 & N & 100.0 & 100.0 & 0.62 & 0.26 & 85.3 & 35.8 \\
 &  & 2 & N & 100.0 & 100.0 & 0.60 & 0.24 & 85.1 & 40.6 \\
 &  & 2 & Y & 100.0 & 100.0 & 0.71 & 0.39 & 83.7 & 33.7 \\ \hline
\multirow{2}{*}{0.6} & \multirow{2}{*}{1.4} & 1 & N & 100.0 & 100.0 & 0.62 & 0.26 & 85.1 & 36.7 \\
 &  & 1 & Y & 100.0 & 100.0 & 0.68 & 0.28 & 86.3 & 35.2 \\ \hline
\multirow{2}{*}{0.9} & \multirow{2}{*}{1.1} & 1 & N & 100.0 & 100.0 & 0.63 & 0.27 & 82.8 & 36.1 \\
 &  & 1 & Y & 100.0 & 100.0 & 0.73 & 0.31 & 79.8 & 32.5 \\ \hline
\end{tabular}
\end{table*}

We explore the advantages of several designs and quantify their influence on the generation results by individually changing or removing them.

\textbf{Effect of the weight coefficient $\alpha$ in Equation (\ref{eq3})}. Equation (\ref{eq3}) converts the discrete data into continuous data by adding noises to support the invertible distribution transformation in flow-based learning. Excessive noises destroy the graph characteristics, yet too small noises influence the graph data continuity. Firstly, TABLE \ref{table4} shows that validity metrics decline with the increase of $\alpha$. This experimental phenomenon is more evident in the edge validity, proving that excessive noise damages the graph information, especially the semantics. Secondly, setting $\alpha$ to 0.9 or 1 performs the best in the MMD metrics. It maintains a good balance between the graph data continuity and the graph characteristics. Lastly, our method achieves the best graph diversity score while setting $\alpha$ to 1, slightly better than setting $\alpha$ to 0.9. However, the graph diversity comes at the cost of reducing the edge validity by a large margin. All things considered, the weight coefficient $\alpha$ is set to 0.9.

\textbf{Effect of the layers of GCN}. GCN learns the node and graph embedding, which is the basis of the Gaussian parameter estimation. In this experiment, we explore the effectiveness of GCN layers on the final results. TABLE \ref{table4} shows that the 3-layer and 5-layer GCN have an evident decline in most metrics compared with the 4-layer one. We also use a 6-layer GCN for this ablation experiment. However, it cannot generate the graphs for many testing scenes, thus not listed in the table. It indicates that the embedding learning of observation data has key effects on the quality of generation results. A GCN with too few layers weakens the model's learning ability and dramatically degrades the graph generation performance. A GCN with too many layers causes the output features over-smoothed \cite{KipfW17, VelickovicCCRLB18, LiHW18, ChenWHDL20}. The indistinguishable features weaken the differences between objects and the differences between spatial relationships, leading to the learning of fuzzy layout experiences. It outputs graphs with low diversities and edge validities for scenes that have never been seen before, and there is a big gap between the generations and real-world layout graphs. In summary, a 4-layer GCN is the best choice.

\textbf{Effect of the cross-entropy loss}. The cross-entropy loss cooperates with the log-likelihood loss to guide the training of 3D-ANF. This experiment discusses whether we can remove the cross-entropy loss and only use the log-likelihood loss to guide the training process. TABLE \ref{table4} shows that removing the cross-entropy loss degrades the generation performance (-19\% edge validity, +0.17 MMD for degrees, +0.11 MMD for clustering coefficients, and -8.8\% for diversity). It proves that the semantic information of the existing instances and relationships in the scene is necessary for the instance-incremental generations of novel graphs.

\subsection{Graph Generation with Constraints}

\begin{figure*}[htbp]
    \centering
    \includegraphics[width=0.8\textwidth]{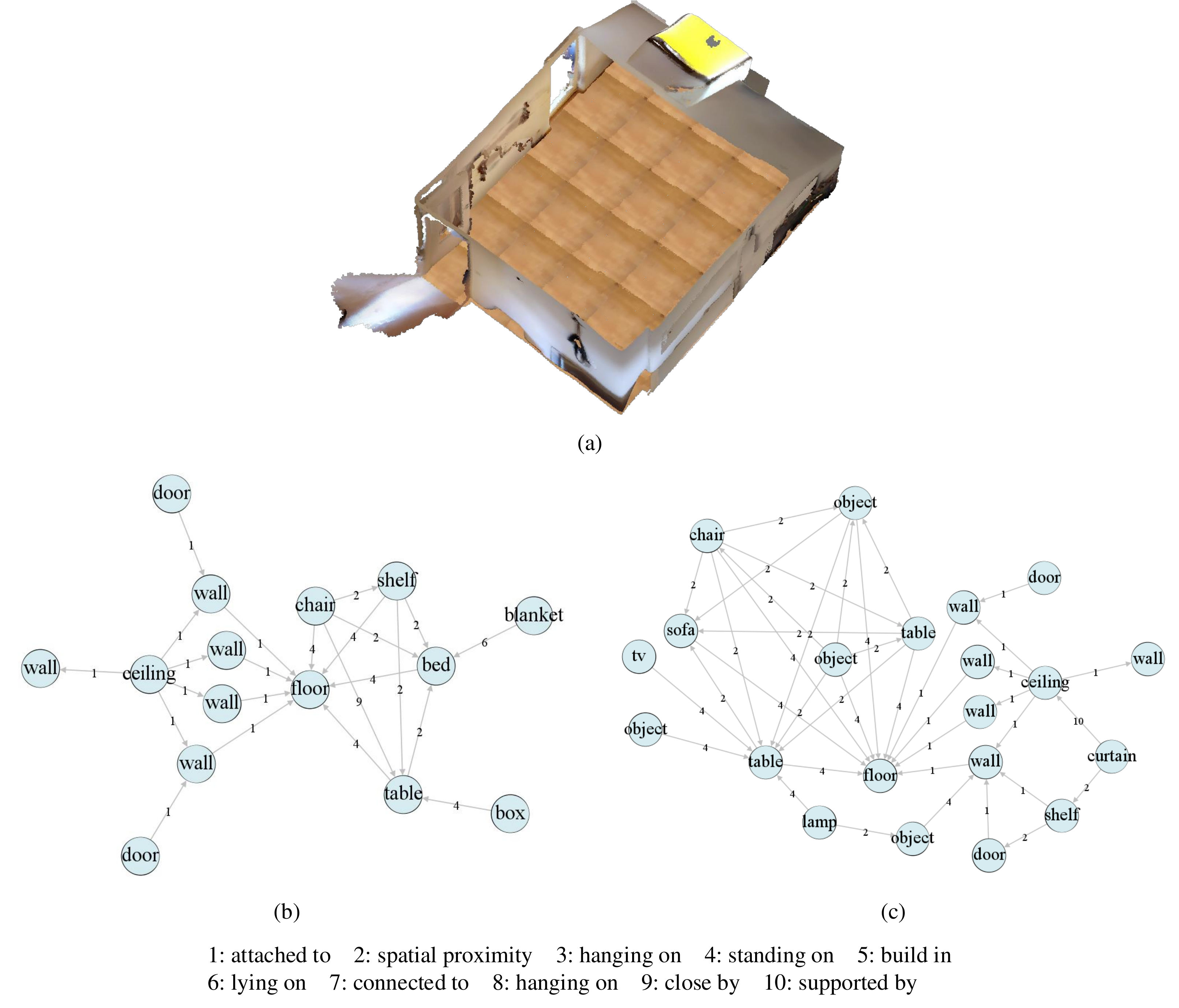}
    \caption{\textbf{Generation visualizations}. (a) A real-world empty living room in the point cloud. (b) Generated graph with constraints. Type of semantic constraint: 1; $\lambda$ of space constraint: 0.8; $\beta$ of space constraint: 1.2. (c) Type of semantic constraint: 2; $\lambda$ of space constraint: 0.6; $\beta$ of space constraint: 1.4.}
    \label{Fig.8}
\end{figure*}

\begin{figure*}[htbp]
    \centering
    \includegraphics[width=0.75\textwidth]{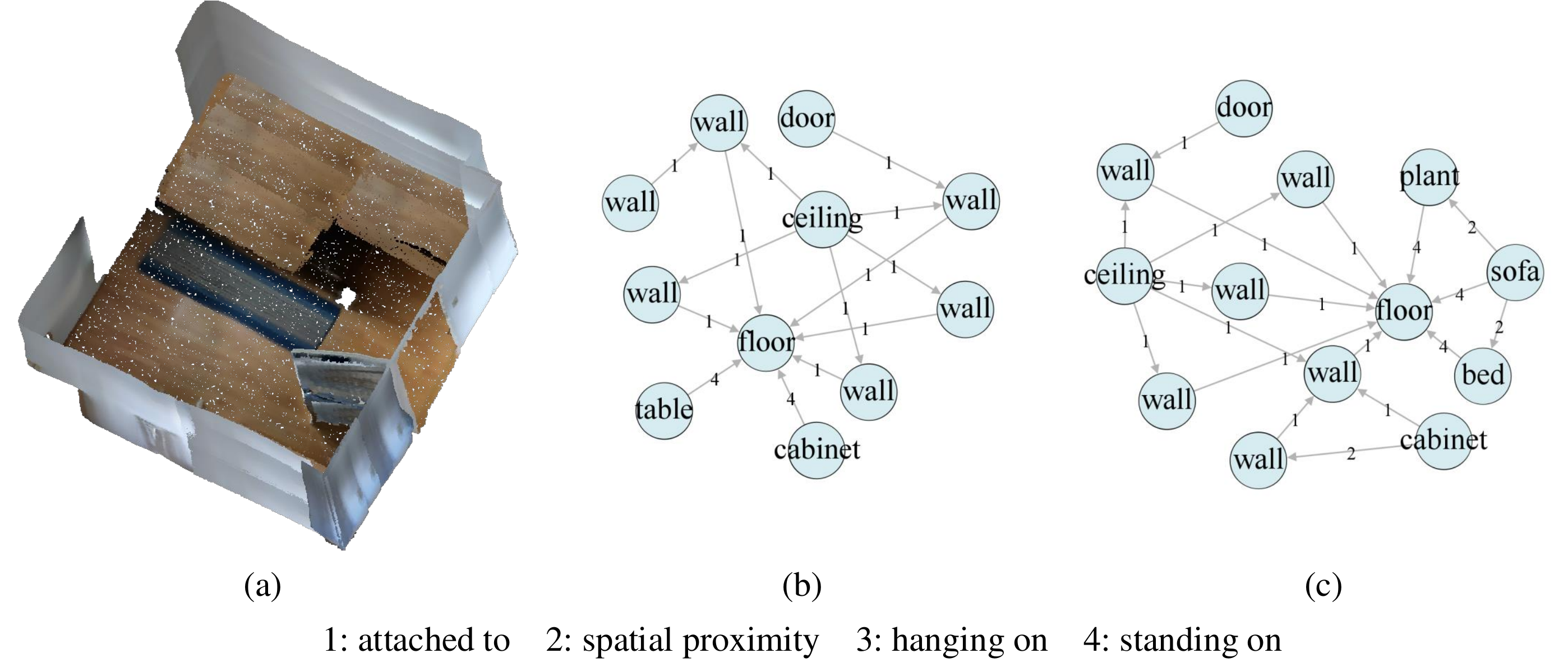}
    \caption{\textbf{Generation visualizations}. (a) A real-world empty bedroom in the point cloud. (b) Generated graph with constraints. Type of semantic constraint: 1; $\lambda$ of space constraint: 0.8; $\beta$ of space constraint: 1.2. (c) Type of semantic constraint: 2; $\lambda$ of space constraint: 0.6; $\beta$ of space constraint: 1.4.}
    \label{Fig.9}
\end{figure*}

\begin{figure*}[htbp]
    \centering
    \includegraphics[width=0.75\textwidth]{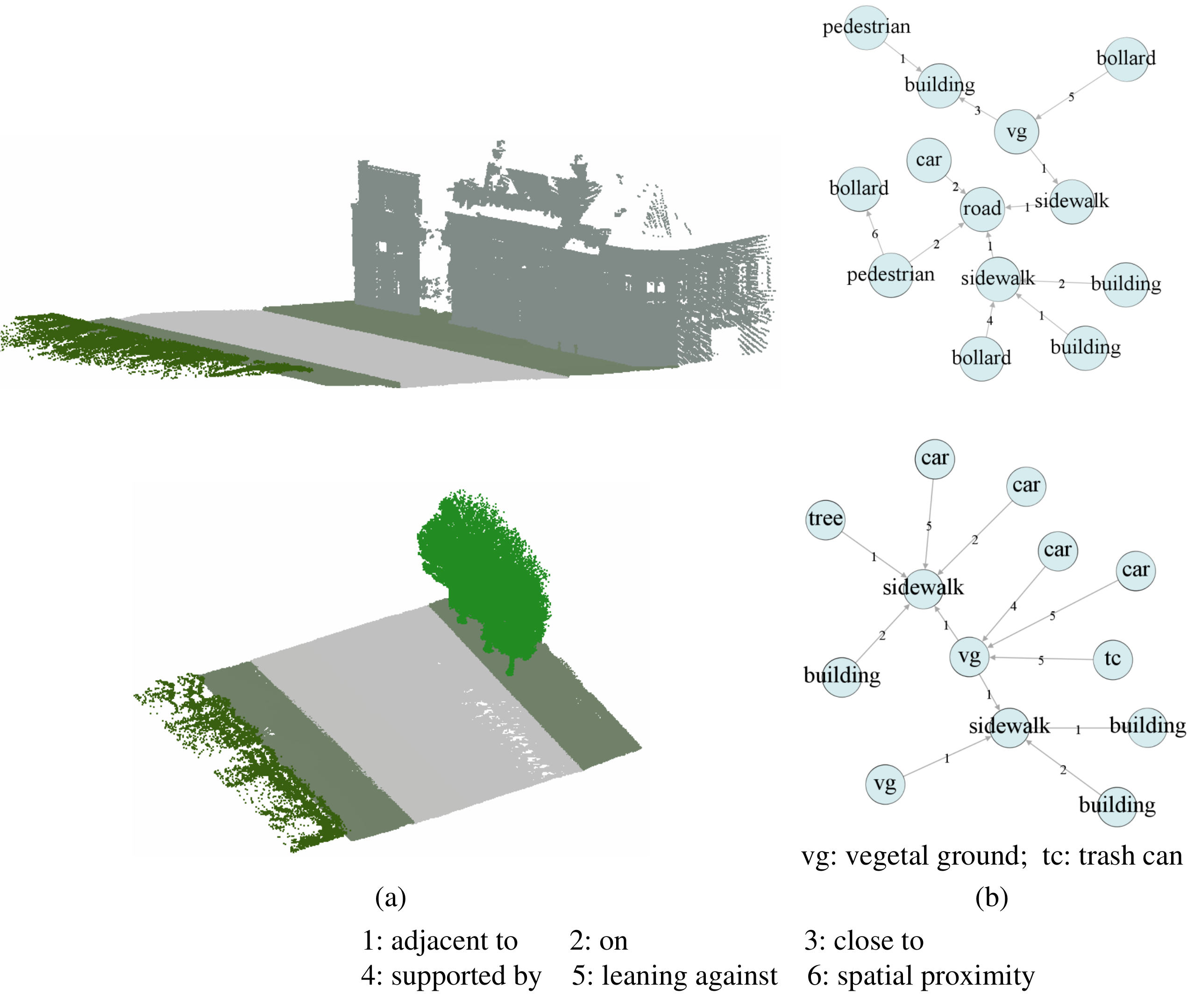}
    \caption{\textbf{Generation visualizations on GPL3D}. (a) Outdoor scenes in Paris-Lille-3D. (b) Generated graphs with semantic constraints and anti-overlapping operations.}
    \label{Fig.10}
\end{figure*}

We add constraints to the generation to make the generated graphs more plausible. We set the lower limit ${L_l} = \lambda  \cdot {L_m}$ and upper limit ${L_u} = \beta  \cdot {L_m}$ of the volume proportion, where ${L_m}$ is the mean value of volume proportion obtained from observation data statistics, $\lambda$ and $\beta$ are individually the weight coefficient for the lower and upper limit. For an empty room with a total volume of ${V_t}$,  the total volume of generated objects should require: ${L_l} \cdot {V_t} \le \sum\limits_{i = 1}^{T_v} {{v_i}}  \le {L_u} \cdot {V_t}$, where ${T_v}$ is the total number of generated objects; and ${v_i}$ is the volume of an object (calculated as mentioned in Section $\uppercase\expandafter{\romannumeral5}$-B).

We also propose two types of semantic constraints. The first one limits the generated objects to furniture, such as \emph{tables} and \emph{sofas}, rather than the existing entities of the empty room, such as \emph{walls} and \emph{floors}. The second one requires the generated objects to be suitable for the room function. The generated edges both needed to conform to the objective reality. Besides, the anti-overlapping operations described in Section $\uppercase\expandafter{\romannumeral5}$-B are selectively used in graph generation with constraints.

TABLE \ref{table5} illustrates the quantitative results with different constraints. By introducing constraints into the generation, the validity of generated nodes and edges all reach 100.0\%. However, the increased values of MMD indicate that the constraints expand the difference between the structures of generated graphs and those of observations. Besides, the decreased graph uniqueness and diversity values prove that these constraints reduce the differences among the generated graphs.

Fig. \ref{Fig.8}, \ref{Fig.9}, and Fig. 4 of the supplementary material-2 show the generated graph with different constraints, considering different real-world rooms in the point cloud as the input. Taking the living room in Fig. \ref{Fig.8} as an example, on the one hand, some unreasonable instances appear in Fig. \ref{Fig.8} (b) without the constraint of the living room. For example, `\emph{a bed in the living room}.' The generated instances in Fig. \ref{Fig.8} (c) are more suitable for a living room by strengthening the constraints on generated objects' categories. On the other hand, space constraints in the generation help control the room's crowding degree. The output layout of our method is very close to that in the real living room.

Fig.2 of the supplementary material-2 illustrates the generation visualization of baseline methods. GraphEBM generates some objects suitable for this room. However, the generated spatial information is insufficient to describe a reasonable layout, making the object arrangement random and messy. SGG-GEMS generated an isolated subgraph without any relationship with the objects in the room, which is inconsistent with the logic of reality.

We construct the corresponding 3D scenes to showcase the generated scene graphs (Supplementary material-2). Note that our generated scene graph provides the category of generated objects with approximate locations for a real-world scene. Lots of other prediction work should be done to achieve the final object arrangement in a real-world scene. They are not the focus of this paper. Thus, we replace them with manual operations to achieve the final object arrangements.

\begin{table*}[htbp]
\centering
\caption{Quantitative results on GPL3D.}
\label{table6}
\begin{tabular}{c|cccccc}
\hline
\multirow{2}{*}{Method} & \multirow{2}{*}{\begin{tabular}[c]{@{}c@{}}Node\\ Validity (\%) $\uparrow$\end{tabular}} & \multirow{2}{*}{\begin{tabular}[c]{@{}c@{}}Edge\\ Validity (\%) $\uparrow$\end{tabular}} & \multicolumn{2}{c}{MMD $\downarrow$} & \multirow{2}{*}{\begin{tabular}[c]{@{}c@{}}Uniqueness\\ \% $\uparrow$\end{tabular}} &
\multirow{2}{*}{\begin{tabular}[c]{@{}c@{}}Diversity\\ \% $\uparrow$\end{tabular}}\\ \cline{4-5}
 &  &  & Degree & Cluster &  \\ \hline
GraphEBM & 68.2 & 39.3 & 1.62 & 1.68 & \textbf{100.0} & \textbf{97.9}\\
GraphRNN & 72.1 & 53.4 & 0.51 & 0.49 & 92.5 & 54.6\\
SGG-GEMS & 81.2 & 56.7 & 0.65 & 0.52 & \textbf{100.0} & 60.8\\
3D-ANF (Ours) & \textbf{88.6} & \textbf{58.9} & \textbf{0.47} & \textbf{0.44} & \textbf{100.0} & 92.1\\ \hline
\end{tabular}
\end{table*}

\subsection{Graph Generation from Outdoor Scenes}

No graphical dataset of outdoor point-cloud-based scenes is available for this task, mainly due to the lack of graph annotations. We have made a new graphical dataset based on the Paris-Lille-3D \cite{roynard2018paris} to address this problem. Paris-Lille-3D is a large-scale urban outdoor point cloud dataset acquired by mobile laser scanners in cities of France, covering more than 2 km of streets. This new dataset is a graphical version of Paris-Lille-3D, which we call Graphical-Paris-Lille-3D (GPL3D). Supplementary material-3 describes the GPL3D in detail. It contains 15 object classes (e.g., \emph{road} and \emph{pedestrian}) and 9 non-empty relationship classes (e.g., \emph{adjacent to} and \emph{close to}). The Lille scenes are for training and Paris for testing.

We conduct comparison experiments on the newly proposed graphical dataset GPL3D, and TABLE \ref{table6} illustrates the experiment results. Our method also achieves the best scores in most evaluation metrics.

The results prove that our method has a better generalization ability than other methods. Specifically, the layout graphs of GPL3D are different from the 3DSSG-O27R16. On the one hand, the graphs have lower feature dimensions due to the fewer label categories. On the other hand, the adjacency matrices are sparse due to the limited annotations of spatial relationships. Our method performs the best on all the datasets, indicating that it can learn the real-world layout experiences from the scenes with different layout graph characteristics. It verifies the generalization ability of our method.

The baseline methods lag behind ours in terms of validities and MMD. Even though the graph complexity of GPL3D decreases a lot, these sequence-based and energy-based methods cannot extract the layout experiences from reality as well as ours due to the limited capability in data distribution modeling from 3D scenes.

The graph diversities of all the methods generally become better. As the first version of GPL3D, the annotations of spatial relationships used for layout learning are not as many as those of 3DSSG-O27R16. A limited number of learning samples leads to weakened restrictions from real-world habits during the generation, resulting in a general increase in graph diversities. Fig. \ref{Fig.10} illustrates the generated graphs with reasonable object layouts. The corresponding 3D scenes of these graphs are in Fig.5 and Fig.6 of supplementary material-2.

\section{Future Work}

The generated scene graphs represent the layout schemes of real-world scenes, recording the object semantics and approximate locations. How to use the layout graph to achieve the final object arrangement in the scene is important. The precise object poses information, including the position and rotation angle, should be given to achieve this goal. Thus, we plan to study object pose prediction with a layout graph as input. It is necessary for the final object arrangement in the scenes.

\section{Conclusions}

This paper explores a new task of instance-incremental graph generation from the real-world point cloud. A framework named 3D-ANF is proposed to solve this conditional generation problem in 3D space. It converts the 3D point cloud into a graph with labeled nodes and semantically meaningful edges through a representation learning module. Besides, Our method learns the real-world experiences from observation data through normalizing flows and autoregressively generates new nodes or edges according to conditions extracted from the current scene. We implement this task on 3DSSG-O27R16 and our proposed GPL3D to verify the effectiveness of our method. Experimental results show that our method generates reliable graphs and outperforms other graph generation methods. It helps to guide the insertion of 3D content into a real indoor scene, adding authenticity to vision-based applications like augmented reality.

\bibliographystyle{IEEEtran}

\end{document}